\documentclass{article}
\usepackage{iclr2027_conference,times}
\usepackage[T1]{fontenc}
\usepackage[utf8]{inputenc}
\usepackage{amsmath,amssymb}
\usepackage{graphicx}
\usepackage{booktabs}
\usepackage{wrapfig}
\usepackage{capt-of}
\usepackage{placeins}
\usepackage{url}
\usepackage[table]{xcolor}
\usepackage{algorithm}
\usepackage{algpseudocode}
\usepackage{listings}
\usepackage[most]{tcolorbox}
\usepackage{hyperref}
\definecolor{promptblue}{HTML}{244858}
\definecolor{promptbackground}{HTML}{F4F8FA}
\tcbset{promptstyle/.style={enhanced,breakable=true,colback=promptbackground,
  colframe=promptblue,boxrule=0.5pt,arc=2mm,left=7pt,right=7pt,
  top=5pt,bottom=5pt,fonttitle=\bfseries\small,coltitle=white,
  listing only,listing options={basicstyle=\ttfamily\fontsize{8}{9.5}\selectfont,
  breaklines=true,breakautoindent=false,columns=fullflexible,
  keepspaces=true,showstringspaces=false}}}
\hypersetup{hidelinks,pdfauthor={Nikita Sorokin, Zoya Volovikova, Alexey Skrynnik, Aleksandr Panov},pdftitle={ZeroHero: Training Small Language Agents from Scratch}}

\title{ZeroHero: Training Small Language Agents\\from Scratch}
\author{Nikita Sorokin\textsuperscript{1}, Zoya Volovikova\textsuperscript{1}, Alexey Skrynnik\textsuperscript{1}, Aleksandr Panov\textsuperscript{2}\\
\normalfont\small \textsuperscript{1} Cognitive AI Systems Lab (cogailab.com)\\
\normalfont\small \textsuperscript{2} Moscow Independent Research Institute of Artificial Intelligence (miriai.org)}
\iclrfinalcopy

\begin{document}
\setlength{\abovecaptionskip}{2pt}
\setlength{\belowcaptionskip}{0pt}
\setlength{\textfloatsep}{10pt plus 1pt minus 2pt}
\setlength{\intextsep}{8pt plus 1pt minus 2pt}
\raggedbottom
\maketitle
\pagestyle{plain}
\begin{abstract}
Large pretrained language models provide capable policies for interactive tasks, but how much model capacity is necessary for a specialized language agent?
We introduce ZeroHero, a framework for training small language agents (SLAs) from scratch. In each environment, a large language model acts as the expert, and we collect its successful trajectories. We pretrain each SLA, which has fewer than 50 million parameters, to predict the expert's responses given the interaction history. We then post-train the agent on its own interactions using proximal policy optimization (PPO). During post-training, reasoning alignment supplements environment rewards with expert supervision on the language the agent generates alongside its actions.
We evaluate the agents in three multistep environments where acting requires both understanding and generating language. In Crafter, agents interpret textual observations and generate reasoning before acting. In WebShop, they translate shopping instructions into search queries. In Alfworld-Dialogue, they ask clarifying questions and use the answers to guide household actions.
The agents approach expert performance on WebShop and surpass their respective demonstration experts on Crafter training rollouts and held-out Alfworld-Dialogue tasks.
An interactive demo is available at \url{https://zoyav.github.io/-ZoyaV-zerohero-demo}.
\end{abstract}

\vspace{-10px}
\begin{figure}[!h]
    \centering
    \includegraphics[width=1\linewidth]{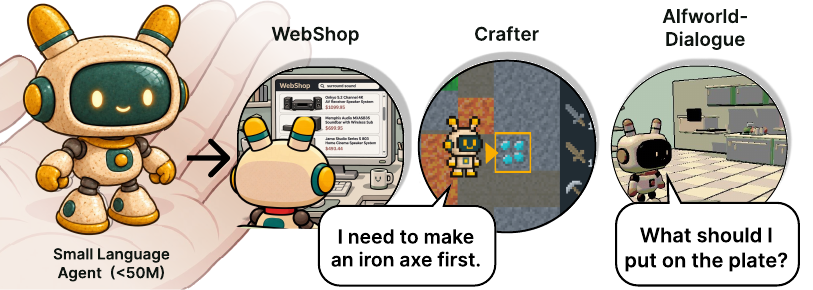}
    \caption{ZeroHero agents use language to reason, search, and ask questions across three interactive environments.}
    \label{fig:zerohero}
\end{figure}

\section{Introduction}
\label{sec:introduction}

Large language models (LLMs) enable a wide range of agentic tasks 
in which achieving a goal requires sequential decision-making, 
and natural language serves as part of the agent's policy. 
An agent can use language to clarify goals through dialogue 
\citep{padmakumar2022teach,gao2022dialfred}, plan and identify 
intermediate goals \citep{yao2023react}, or directly execute 
actions, such as formulating search queries. However, the 
applicability of LLMs is limited by their computational 
requirements. Limited Internet connectivity or privacy 
requirements may prevent the use of cloud-based services, 
while running large models locally requires substantial 
computational resources and memory. These limitations 
motivate the investigation of compact language models 
and training methods for specialized agentic tasks.

Previous studies have demonstrated that Transformer-based 
models with only a few million parameters can address 
a wide range of specialized tasks, from natural language 
generation to sequential decision-making. In TinyStories, 
\citet{eldan2023tinystories} demonstrate that language 
models with fewer than 10 million parameters can generate 
coherent short stories. In contrast, MAPF-GPT 
\citep{andreychuk2025mapfgpt} employs models with 2 and 
6 million parameters to solve multi-agent pathfinding 
tasks, while DART \citep{agarwal2024dart} employs compact 
Transformer-based models for world modeling and policy 
learning through reinforcement learning in interactive 
environments. In the latter two studies, the agents' 
policies do not rely on natural language. This raises 
the question of whether models of comparable size can 
combine natural language understanding and generation 
with sequential decision-making in interactive environments.

In this work, we investigate whether a language model 
with fewer than 10 million parameters can be trained 
from scratch to solve multi-step tasks in which natural 
language understanding and generation are integral 
components of the agent's policy. We consider three 
environments in which language serves different purposes. 
In \textit{Alfworld-Dialogue}, the agent engages in dialogue 
to clarify household task requirements and obtain 
information necessary for task completion. In 
\textit{Crafter}, the agent reasons about its observations 
and maintains textual notes that can inform subsequent 
decisions. In \textit{WebShop}, the agent follows user 
instructions by sequentially invoking tools to search 
for and select products. For each environment, we train a separate agent in two 
stages. First, the language model is trained from scratch 
on demonstrations generated by an expert LLM. Second, 
the agent interacts with the environment and refines 
its policy through reinforcement learning. Our experimental 
results demonstrate that language agents with fewer than 
10 million parameters can solve multi-step tasks requiring 
the use of natural language and, in some cases, achieve 
higher success rates than the expert models with up to 
30 billion parameters that provided the demonstrations.
\section{Related Work}

\textbf{Small language model training from scratch.}
Recent work shows that compact language models can achieve strong task-specific performance through better use of data, architecture, and supervision~\citep{eldan2023tinystories,gunasekar2023textbooks,zhang2024tinyllama,hu2024minicpm,liu2024mobilellm,mehta2024openelm}. Larger LLMs can also serve as teachers: \citet{hsieh2023distilling} distill teacher-generated rationales into smaller models, while \citet{he2026learning} study how expert supervision and RL can be combined across different stages of SLM training. We extend this direction to interactive decision making, asking whether a randomly initialized SLA can learn both action and dialogue behavior and eventually surpass its expert.

\paragraph{RL fine-tuning of LLMs and VLMs.}
Pretrained LLMs and VLMs provide useful priors for decision making, motivating their use as policy backbones in multi-turn environments~\citep{tan2024twosome,wang2025ragen,wang2025vagen}. However, RL fine-tuning remains difficult due to long-horizon credit assignment, sparse rewards, exploration, and training instability~\citep{wang2025ragen,wang2026ragen2}. GiGPO specifically addresses credit assignment by introducing relative advantages at both the episode and step levels for multi-turn LLM agents~\citep{feng2025gigpo}. Moreover, the benefit of pretraining is task-dependent: Odysseus reports improved sample efficiency from VLM pretraining~\citep{shi2026odysseus}, while specialized RL methods can outperform LLM-based policies in low-level control tasks~\citep{hatgiskessell2026promptpo}. We instead use a large model as an expert and train a smaller policy specialized for the target environment.

\paragraph{Embodied dialogue systems and environments.}
Embodied language environments combine language with environment interaction: ALFWorld links text interaction to embodied household tasks~\citep{shridhar2021alfworld}, while TEACh, DialFRED, and IGLU study task completion through dialogue, clarification, or human guidance~\citep{padmakumar2022teach,gao2022dialfred,kiseleva2022interactive}. Related systems such as MindAgent and LEGENT further explore multimodal and language-mediated coordination in embodied settings~\citep{cheng2024legent}. Our setting separates a compact executor from a privileged LLM oracle with access to hidden environment state, allowing the executor to learn not only how to act, but also when and what to ask.

\section{Method}
\label{sec:method}
\begin{figure}[t]
\centering
\includegraphics[width=\linewidth]{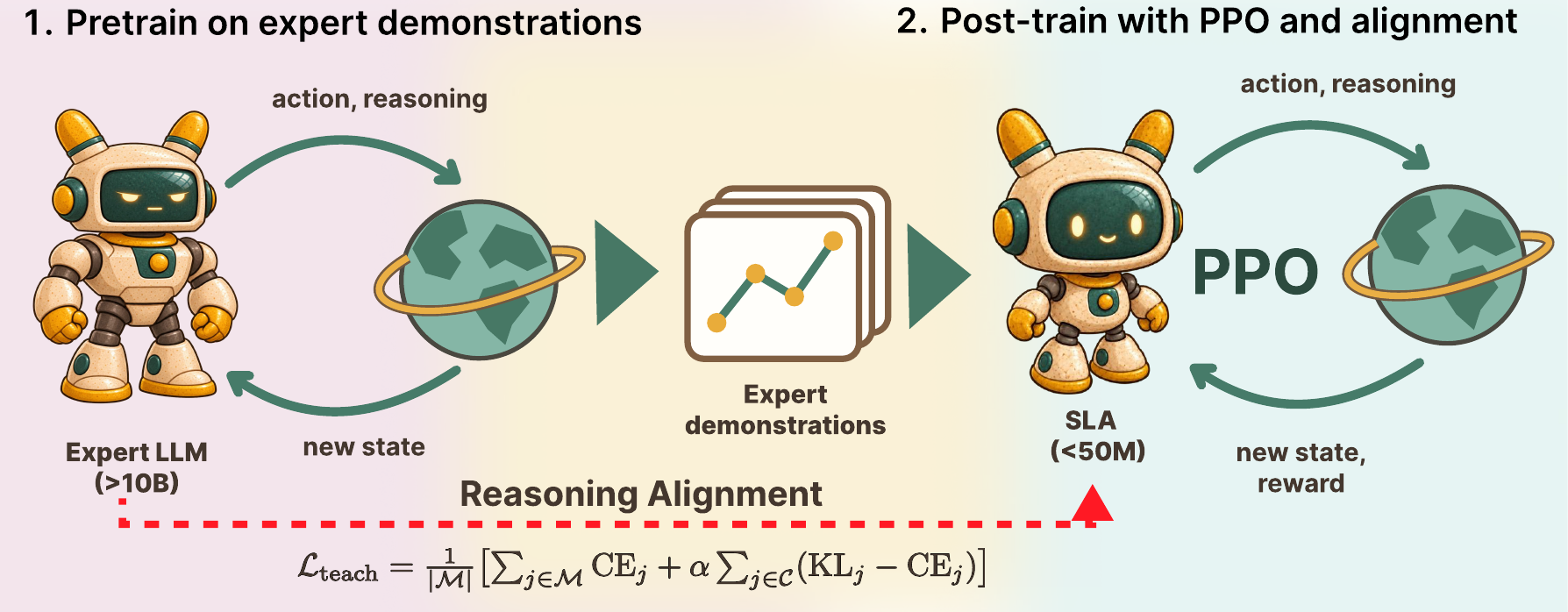}
\caption{ZeroHero training. The student first learns from successful expert trajectories. During post-training, PPO updates the student using environment rewards, while reasoning alignment uses expert supervision on language tokens.}
\label{fig:method-overview}
\end{figure}

ZeroHero trains a language agent with up to 50 million parameters from scratch in two stages (Figure~\ref{fig:method-overview}). We first run a larger language model as an expert in the task environment and collect its successful interaction trajectories. We use these trajectories for supervised pretraining, teaching the student to imitate the expert responses given the recorded interaction histories (Section~\ref{sec:agent-pretrain}). During post-training, the student interacts with the same environment and updates its policy using the rewards it receives (Section~\ref{sec:agent-posttrain}). We augment this stage with reasoning alignment, which uses expert responses to supervise the language the student generates alongside its actions (Section~\ref{sec:reasoning-alignment}).

\subsection{Small Language Agent Pretrain}
\label{sec:agent-pretrain}
A larger language model interacts with the environment to generate demonstrations. From each successful episode, we pair the history available at each decision with the expert response that followed.

At decision $t$, the agent receives a history $h_t$ containing its instruction, current observation, and the interaction context available through the task interface. It generates a response $y_t=(y_{t,1},\ldots,y_{t,L_t})$, from which an action is extracted and executed. Depending on the environment, a response may also contain reasoning, a question, or an updated memory. We denote the autoregressive student by $\pi_\theta(y\mid h)=\prod_i\pi_\theta(y_i\mid h,y_{<i})$.

Expert trajectories provide a dataset $\mathcal D_E=\{(h,y^*)\}$ of histories and target responses. We initialize $\theta$ randomly and minimize the response-token cross-entropy
\begin{equation}
\mathcal L_{\mathrm{SFT}}(\theta)
=-\frac{1}{N_E}\sum_{(h,y^*)\in\mathcal B_E}\sum_{i=1}^{|y^*|}
\log\pi_\theta(y_i^*\mid h,y^*_{<i}),
\qquad N_E=\sum_{(h,y^*)\in\mathcal B_E}|y^*|.
\label{eq:zh-sft}
\end{equation}
Here $\mathcal B_E$ is a demonstration minibatch. Targets include the response structure, action, and end-of-sequence token; prompt and padding tokens are excluded. The resulting policy $\pi_0$ initializes reinforcement learning and is retained as a frozen language reference.

\subsection{Small Language Agent Postrain}
\label{sec:agent-posttrain}
The student collects trajectories under a behavior policy $\pi_{\mathrm{old}}$. Each generated response constitutes one environment decision, with reward $r_t$, a return target $\widehat R_t$, and an estimated advantage $\widehat A_t$. We write the PPO score and surrogate ratio as
\begin{equation}
\ell_\theta(h_t,y_t)=\frac{1}{c_t}\sum_{i=1}^{L_t}
\log\pi_\theta(y_{t,i}\mid h_t,y_{t,<i}),
\qquad
\rho_t(\theta)=\exp\!\left(\ell_\theta(h_t,y_t)-\ell_{\mathrm{old}}(h_t,y_t)\right).
\label{eq:zh-ratio}
\end{equation}
The normalization $c_t$ determines whether the score sums or averages the token log probabilities. The behavior and updated policies use the same normalization. With length averaging, $\rho_t$ is a surrogate for the sequence likelihood ratio.

Following PPO~\citep{schulman2017ppo}, the actor minimizes the clipped surrogate
\begin{equation}
\mathcal L_{\mathrm{PPO}}(\theta)
=-\frac{1}{|\mathcal B|}\sum_{t\in\mathcal B} w_t\,
\min\!\left[\rho_t(\theta)\widehat A_t,
\operatorname{clip}(\rho_t(\theta),1-\epsilon,1+\epsilon)\widehat A_t\right],
\label{eq:zh-ppo}
\end{equation}
Here $w_t$ weights the contribution of a response. The critic estimates the return from the history $h_t$ and is trained against $\widehat R_t$ with loss $\mathcal L_V$.

\subsection{Reasoning Alignment}
\label{sec:reasoning-alignment}
Environment rewards evaluate the consequences of actions but do not directly supervise the language accompanying them. PPO-RA therefore combines a frozen-policy alignment term during PPO with periodic teacher-supervised updates. The SFT reference preserves the learned language prior; a separate copy of the current actor regularizes action predictions during each auxiliary phase. 

\paragraph{Language alignment during PPO.}
Let $m_{t,i}\in\{0,1\}$ select language tokens in a sampled response. The response format determines which positions enter the mask. With $p_{\theta,t,i}=\pi_\theta(\cdot\mid h_t,y_{t,<i})$, the alignment loss is
\begin{equation}
\mathcal L_{\mathrm{anchor}}(\theta)
=\frac{\sum_{t\in\mathcal B}\sum_i m_{t,i}\,
D_{\mathrm{KL}}(p_{0,t,i}\,\|\,p_{\theta,t,i})}
{\sum_{t\in\mathcal B}\sum_i m_{t,i}},
\qquad
\mathcal L_{\mathrm{actor}}=\mathcal L_{\mathrm{PPO}}+\beta\mathcal L_{\mathrm{anchor}}.
\label{eq:zh-anchor}
\end{equation}
Both distributions are evaluated on the same sampled prefixes.

\paragraph{Teacher-supervised auxiliary updates.}
At scheduled intervals, PPO-RA revisits expert responses, using separate policy and auxiliary phases as in Phasic Policy Gradient (PPG)~\citep{cobbe2021ppg}. Let $\pi_c$ be a frozen copy of the actor at the start of this auxiliary phase. For a teacher minibatch, let $\mathcal M$ contain language positions and $\mathcal A$ contain action positions. A subset $\mathcal C\subseteq\mathcal M$ has usable teacher-distribution targets. At position $j$, define $u_j=-\log\pi_\theta(y_j^*\mid h,y_{<j}^*)$ and let $d_j^T=D_{\mathrm{KL}}(q_j^T\,\|\,\overline p_{\theta,j})$, where $q_j^T$ is the projected teacher target and $\overline p_{\theta,j}$ aggregates student probabilities into the same explicit-token and residual categories. The auxiliary language loss is
\begin{equation}
\mathcal L_{\mathrm{teach}}(\theta)
=\frac{1}{|\mathcal M|}
\left[\sum_{j\in\mathcal M\setminus\mathcal C}u_j
+\sum_{j\in\mathcal C}\big((1-\alpha)u_j+\alpha d_j^T\big)\right].
\label{eq:zh-teach}
\end{equation}
For heterogeneous tokenizers, byte-prefix marginalization~\citep{wang2026bpm} retains unmatched mass in a residual bucket. These targets approximate cross-tokenizer distillation. Positions without eligible targets retain full cross-entropy supervision, including structure and stopping tokens where applicable.

To limit interference with the current action policy, the auxiliary objective includes
\begin{equation}
\mathcal L_{\mathrm{aux}}(\theta)
=\mathcal L_{\mathrm{teach}}(\theta)
+\lambda_a\frac{1}{|\mathcal A|}\sum_{j\in\mathcal A}
D_{\mathrm{KL}}(p_{c,j}\,\|\,p_{\theta,j}),
\label{eq:zh-aux}
\end{equation}
with the action term omitted when $\mathcal A$ is empty. Auxiliary updates train the actor while keeping the critic fixed.

\begin{algorithm}[H]
\caption{ZeroHero: expert pretraining followed by PPO-RA}
\label{alg:zh-pipeline}
\small
\begin{algorithmic}[1]
\Require Expert trajectories $\mathcal D_E$, teacher targets, interaction budget, auxiliary schedule
\State Randomly initialize student $\pi_\theta$
\State Train on $\mathcal D_E$ with $\mathcal L_{\mathrm{SFT}}$ (Eq.~\ref{eq:zh-sft}); freeze the resulting reference $\pi_0$
\State Initialize critic and separate PPO and auxiliary optimizers
\While{the interaction budget is not exhausted}
  \State Freeze behavior policy $\pi_{\mathrm{old}}\gets\pi_\theta$; collect trajectories $\mathcal B$
  \State Compute $\widehat R_t$, $\widehat A_t$, and behavior scores using Eq.~\ref{eq:zh-ratio}
  \For{each PPO minibatch}
    \State Update the critic with $\mathcal L_V$ and the actor with $\mathcal L_{\mathrm{actor}}$ (Eq.~\ref{eq:zh-anchor})
  \EndFor
  \If{an auxiliary phase is due}
    \State Freeze $\pi_c\gets\pi_\theta$ for action regularization
    \For{each scheduled auxiliary step}
      \State Sample a minibatch of expert responses and their teacher targets
      \State Update the actor with $\mathcal L_{\mathrm{aux}}$ (Eq.~\ref{eq:zh-aux}), keeping the critic fixed
    \EndFor
  \EndIf
\EndWhile
\end{algorithmic}
\end{algorithm}

Plain PPO disables the anchor term and auxiliary phases. A comparison with plain PPO measures the joint effect of the anchor and auxiliary updates.

\section{Experiments}
\label{sec:setup}
We ask three questions about training SLAs.
\begin{enumerate}
\item Can a model with up to 50 million parameters learn a policy that requires both multistep decision-making and language generation from expert demonstrations? How many expert transitions does it need to learn such a policy?
\item Can reinforcement learning improve a pretrained agent's task performance and enable it to surpass its demonstration expert in the same domain?
\item What quality of reasoning and language do the agents exhibit after pretraining and RL, and how does reasoning alignment affect that quality?
\end{enumerate}
Section~\ref{sec:experimental-setup} describes the setup used to investigate these questions.

\subsection{Experimental Setup}
\label{sec:experimental-setup}
\paragraph{Environments.}
We focus on three environments in which language supports different parts of the policy: reasoning about an action, constructing search queries, and eliciting missing task information. 

The first environment, \textit{Crafter}~\citep{hafner2022crafter}, is a Minecraft-like open world. The agent explores, manages health, hunger, and thirst, defends itself against mobs, and works toward achievements with resource and tool prerequisites. It generates reasoning before choosing a low-level action, such as moving in a cardinal direction, gathering resources with \texttt{DO}, placing a crafting table with \texttt{PLACE\_TABLE}, or crafting a pickaxe. Crafter has 22 achievements. One simplified path to a diamond is \mbox{\raisebox{-0.18em}{\includegraphics[height=0.95em]{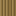}} $\to$ \raisebox{-0.18em}{\includegraphics[height=0.95em]{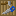}} $\to$ \raisebox{-0.18em}{\includegraphics[height=0.95em]{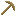}} $\to$ \raisebox{-0.18em}{\includegraphics[height=0.95em]{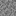}} $\to$ \raisebox{-0.18em}{\includegraphics[height=0.95em]{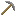}} $\to$ \raisebox{-0.18em}{\includegraphics[height=0.95em]{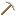}} $\to$ \raisebox{-0.18em}{\includegraphics[height=0.95em]{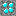}}}.

In the second environment, \textit{WebShop}~\citep{yao2022webshop}, the agent must purchase a product that satisfies a natural-language request. It writes and revises search queries, inspects products, and selects options before buying. The generated queries determine which products the agent can discover. We use a synthetic catalog of 1,000 products.

In the third environment, \textit{Alfworld-Dialogue}, we retain ALFWorld's six household task families and physical success
predicates \citep{shridhar2021alfworld}: placement, examination in light,
cleaning, heating, cooling, and two-object placement. Instead of receiving
the complete goal, the agent receives a request with missing requirements,
such as the destination, treatment, or quantity. The omitted fields are
fixed for each task, allowing policies to be compared on the same requests.
The agent can issue a physical command or a free-text \texttt{ask} question.
An oracle backed by Qwen3-4B answers from a private task/state FAQ. The
original full goal is absent from the agent's observations and history.

\paragraph{Architecture.}
\label{sec:architectures}
The student is an autoregressive decoder-only Transformer. Its input is a tokenized interaction history. A causal decoder predicts the next response token from this history and the response prefix. The same policy generates language and executable actions. During PPO, a critic estimates returns for training; the actor alone generates responses at execution time.

We use a Qwen-like decoder-only Transformer in three nominal size classes: 5M, 10M, and 50M parameters. All three use rotary positional embeddings, RMSNorm, SwiGLU, grouped-query attention, and tied input and output embeddings.

\paragraph{Experts and expert demonstrations.}
We use a task-specific expert to demonstrate how language supports decisions in each environment. In Crafter, Qwen3.8-27B provides reasoning and low-level actions. In WebShop, a GiGPO-trained Qwen2.5-7B provides search and purchase trajectories using a policy already trained on the shopping task. In Alfworld-Dialogue, Gemma-4-26B-A4B-it receives the public observation and oracle replies with additional instructions for deciding when to ask or act. Experts can receive expanded prompts: WebShop provides an explicit action menu, and the Crafter alignment teacher receives additional guidance on reasoning about action prerequisites. Appendix~\ref{app:prompts} provides the expert and SLA prompt templates.

Each transition contributes one recorded decision to the corpus. The largest training corpora contain 85,074 transitions across 384 worlds for Crafter and 42,579 transitions across 8,136 instructions for WebShop. After screening responses and oracle replies, Alfworld-Dialogue contributes 63,379 training transitions from 5,224 successful trajectories spanning 2,510 games. Smaller corpora form the data-scaling sweep. Validation holds out distinct ALFWorld games.

\paragraph{Pretraining setup.}
Across all three environments, we train randomly initialized agents with response-only cross-entropy, including the end-of-sequence token, conditioned on recorded interaction histories. We use AdamW and task-specific byte-level BPE tokenizers. Targets contain reasoning and actions in Crafter and WebShop, and commands or clarification questions in Alfworld-Dialogue. The data-scaling experiments compare the same nominal model sizes (5M, 10M, and 50M) over 24 epochs, selecting checkpoints by development-set task performance.

\paragraph{Post-training setup.}
We initialize post-training from pretrained SLAs: Crafter uses a 5M agent at epoch 12, trained on 20,949 transitions from 96 worlds; WebShop uses a 10M agent at epoch 4, trained on 22,005 transitions from 4,068 instructions; Alfworld-Dialogue uses a 5M agent at epoch 12, trained on 32,578 transitions from 2,612 trajectories. Within each environment, PPO and PPO-RA start from the same checkpoint and share rollout and PPO settings; their learning curves are compared at matched interaction budgets.

For full-response PPO, $c_t=w_t=1$ in Eqs.~\ref{eq:zh-ratio} and~\ref{eq:zh-ppo}. The long-response formulation uses $c_t=L_t$ and $w_t=L_t/\overline L_{\mathcal B}$, where $\overline L_{\mathcal B}$ is the minibatch mean response length. Scalar critics use half the mean squared return error. Distributional critics use cross-entropy against the return target projected onto fixed value bins. Paired PPO and PPO-RA configurations share the return estimator, score convention, critic parameterization, and base entropy term.

PPO-RA adds a frozen-reference language anchor and periodic teacher-supervised auxiliary updates with a separate optimizer. The alignment mask follows the response format, covering command and question tokens in Alfworld-Dialogue. There, the frozen-reference and action-regularization divergences use top-token probabilities and a pooled residual category.

\par
Success rate (SR) measures task completion according to the environment
predicate. We use validation tasks to track learning and test tasks for held-out assessment. Test sets
include 134 Alfworld-Dialogue tasks across six families, with matching tasks and hidden goals across methods. Alfworld-Dialogue uses 60 validation tasks. Each episode permits 60 decisions, 50 physical steps, and eight questions.
Decoding settings appear in Appendix~\ref{app:prompts}.
\par

\begin{figure}[!t]
\centering
\includegraphics[width=\linewidth]{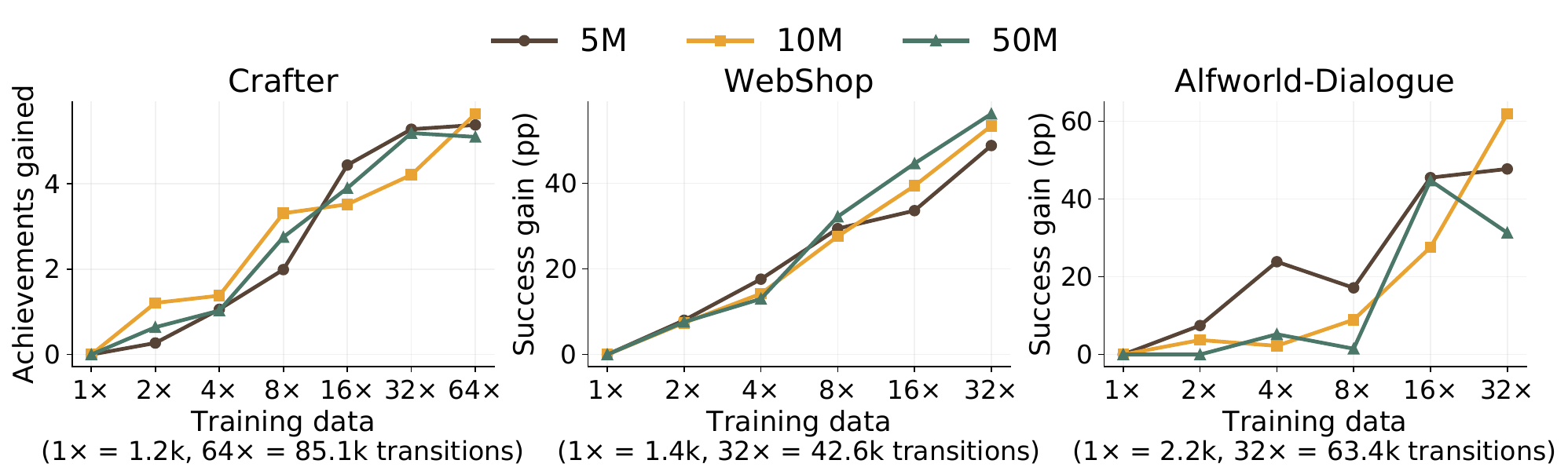}
\caption{Task performance gains for 5M, 10M, and 50M agents relative to their smallest expert corpus (1$\times$).}
\label{fig:rq1-data-scaling}
\par\vspace{4pt}
\includegraphics[width=\linewidth]{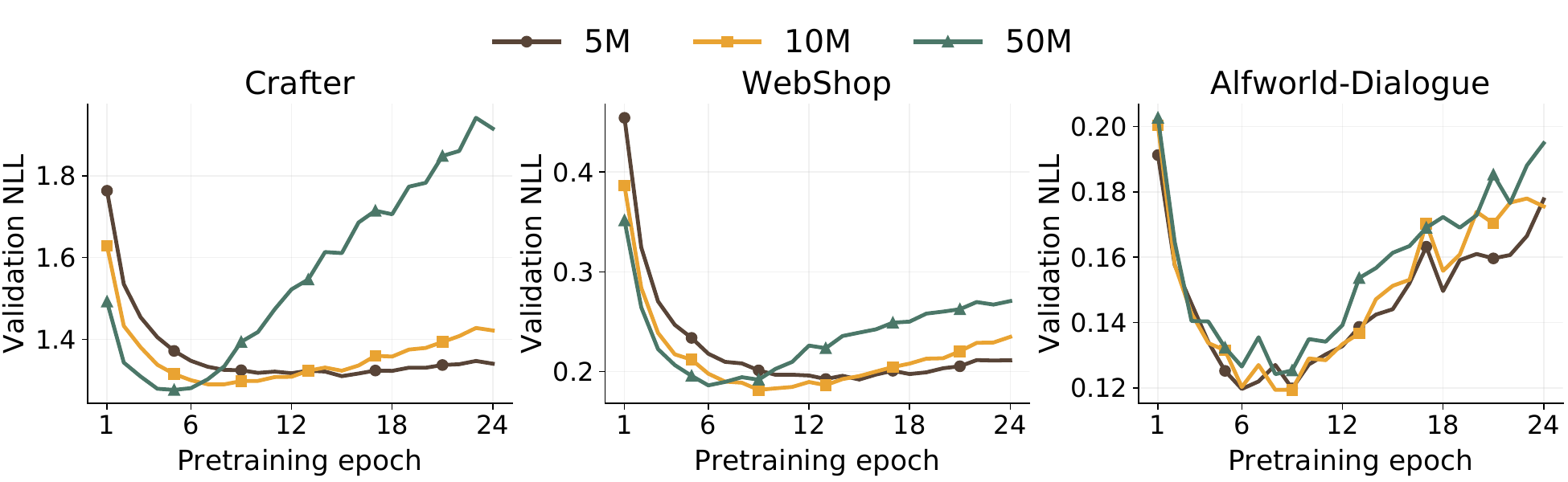}
\caption{Response-token validation loss over 24 epochs on the largest expert corpus in each environment.}
\label{fig:rq1-validation-loss}
\end{figure}

\subsection{RQ1: Learning from Expert Trajectories}
\label{sec:pretraining}
We began by asking how much an agent could learn from demonstrations alone. For each environment, we pretrained agents of three sizes (5M, 10M, and 50M parameters) on progressively larger corpora, doubling the nominal data budget at each level. Even on the smallest corpus, Crafter agents earned 3.10--3.53 of the 22 achievements per episode. Success was rarer in the other environments, but the agents already completed some tasks: 0.8--3.4\% of WebShop tasks and 3.7--9.7\% of Alfworld-Dialogue tasks on their respective smallest corpora.

As we increased the number of demonstrations, all three model sizes improved in Crafter and WebShop, although successive doublings produced uneven gains (Figure~\ref{fig:rq1-data-scaling}). In Crafter, the models reached nearly the same performance on the largest corpus, averaging 8.61--8.74 achievements per episode. Model size mattered more in WebShop. On the largest corpus, increasing capacity from 5M to 10M raised success from 49.6\% to 56.8\%; increasing it again to 50M added only 1.2 percentage points.

In Alfworld-Dialogue, the agent must decide when a household instruction is ambiguous and what question would resolve it. Here, additional demonstrations produced less consistent gains. The 10M agent completed 5.2\% of tasks on the smallest corpus and 7.5\% with 8,020 transitions, but reached 67.2\% success on the largest. On this largest corpus, it outperformed both the 5M agent (51.5\%) and the 50M agent (41.0\%). We also examined how well each model predicted held-out expert responses (Figure~\ref{fig:rq1-validation-loss}).

\subsection{RQ2: Reinforcement Learning After Pretraining}
\label{sec:posttraining-results}
\begin{figure}[!t]
\centering
\includegraphics[width=\linewidth,trim=0 8bp 0 0,clip]{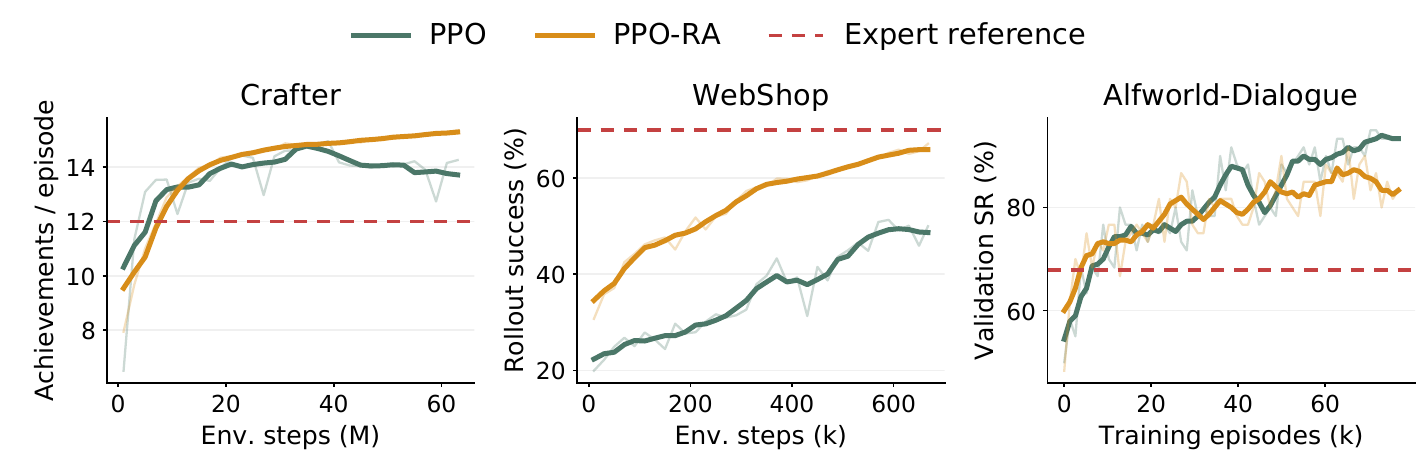}
\caption{Post-training performance of PPO and PPO-RA at matched interaction budgets. Red dashes mark expert references. Crafter and WebShop rollout measurements are grouped into bins of two million and 20,000 environment steps, respectively. Faint lines show unsmoothed values and solid lines show centered five-point means.}
\label{fig:rq2-posttraining}
\end{figure}

We next asked whether a student that had learned to imitate the expert could improve by acting in the environment itself. We continued training pretrained agents with PPO and tracked when their performance reached the expert reference (Figure~\ref{fig:rq2-posttraining}). In Crafter, the PPO agent first exceeded the reference of 12 achievements per episode in the interval from 4 to 6 million environment steps, averaging 13.1 achievements across the logged rollout measurements in that interval. By the final shared bin it averaged 14.3. WebShop showed a different trajectory: success improved with interaction, but even the stronger aligned run finished at 67.2\%, below the expert reference of 70\%. Both comparisons describe performance during training.

Alfworld-Dialogue allowed us to test whether the improvement carried over to unseen household tasks. Starting from a checkpoint that already solved half the validation tasks, PPO reached 93.3\% validation success after 600 cycles.

\subsection{RQ3: Does Reasoning Alignment Help?}
\label{sec:reasoning-results}
Having established what PPO could learn from environment rewards, we asked what changed when we also supplied expert supervision during post-training. We compared plain PPO with PPO-RA at matched interaction budgets (Figure~\ref{fig:rq2-posttraining}). Alignment made the largest difference in WebShop, where the aligned agent completed 67.2\% of shopping tasks in the final training bin and plain PPO completed 50.0\%. In Crafter, alignment added about one achievement per episode (15.4 versus 14.3). The same intervention had a different outcome in Alfworld-Dialogue: after 600 cycles the aligned run solved 83.3\% of validation tasks, ten percentage points below plain PPO.

\begin{wraptable}{r}{0.54\textwidth}
\vspace{-8pt}
\centering\small
\setlength{\tabcolsep}{3pt}
\begin{tabular*}{\linewidth}{@{\extracolsep{\fill}}lrrr@{}}
\toprule
Policy & \shortstack{RL\\steps} & \shortstack{Repeated\\\texttt{ACTION:} (\%)} & \shortstack{Repeat-4\\(\%)} \\
\midrule
Pretrained & 0 & 0.00 & 0.61 \\
PPO & 29.38M & 71.88 & 19.91 \\
PPO-RA & 15.86M & 3.13 & 0.51 \\
\bottomrule
\end{tabular*}
\caption{Crafter response repetition on the same 64 prompts. Repeated \texttt{ACTION:} is the percentage of responses with multiple action markers. Repeat-4 is the mean fraction of duplicate word 4-grams in the pre-action text. Lower values indicate less repetition; neither metric measures semantic correctness.}
\label{tab:crafter-repetition}
\end{wraptable}
We also examined response repetition on 64 fixed Crafter prompts (Table~\ref{tab:crafter-repetition}). Plain PPO frequently repeated the \texttt{ACTION:} marker and word sequences, whereas PPO-RA remained close to the pretrained baseline. The checkpoints have different training budgets, so this probe describes their outputs rather than a matched-budget effect of alignment. Whether alignment improves the semantic quality of explanations or questions remains open, for example whether a clarification question asks for the information missing from a household instruction.

\begin{figure}[!t]
    \centering
    \includegraphics[width=\linewidth]{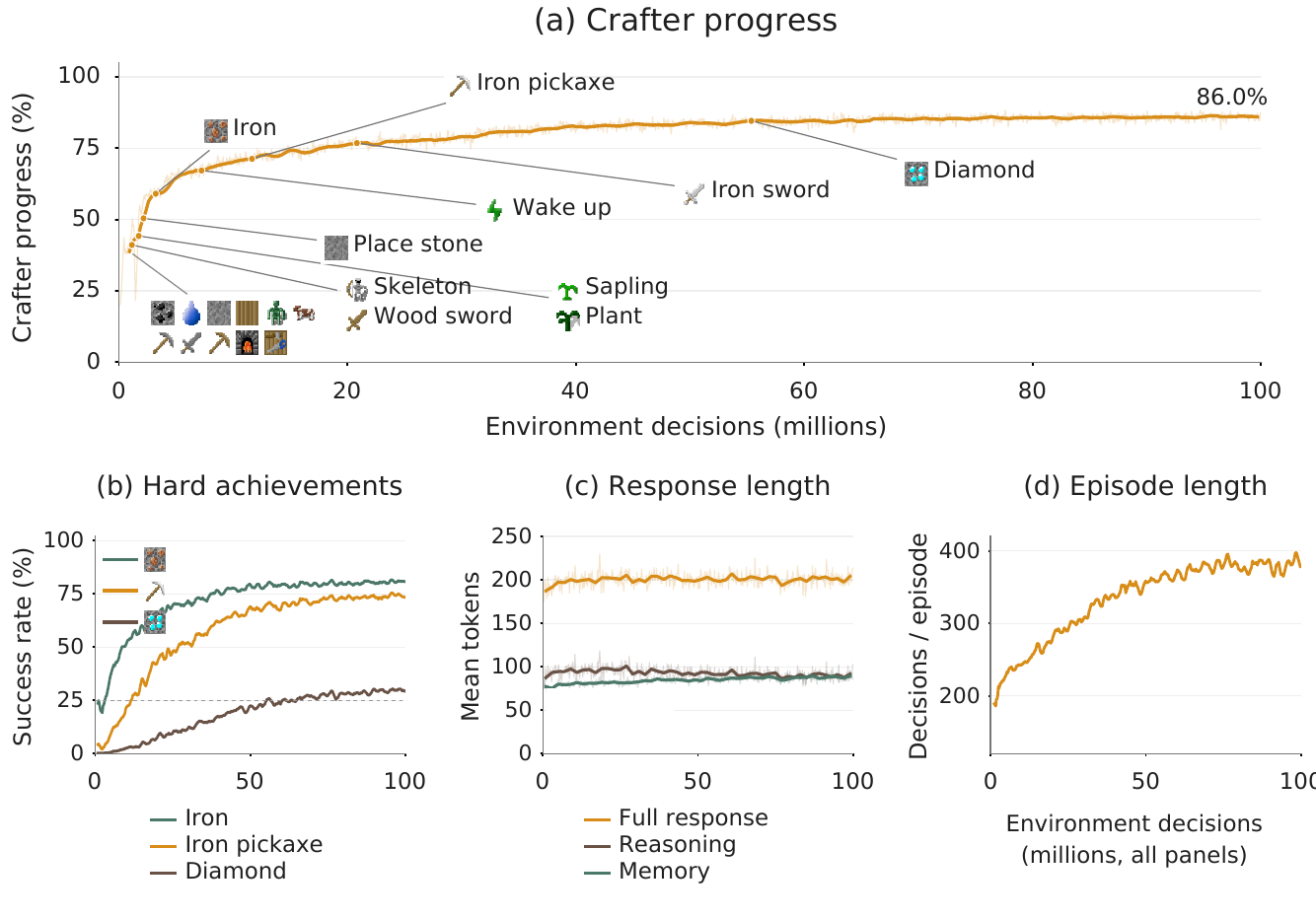}
    \caption{Visual Crafter with PPO-RA through 100M decisions.
    (a) Crafter Progress ($100\,\overline{A}/22$):
    Icons mark first observed achievement success above 25\%;
    the initial group contains skills already above threshold at the start of the plotted curve.
    (b) Iron, iron-pickaxe, and diamond success rates.
    (c) Full-response, reasoning, and memory lengths: 
    Full responses include actions and formatting; block lengths exclude tags.
    (d) Mean agent decisions per episode.}
    \label{fig:visual-crafter-learning}
\end{figure}

\section{Scaling Visual Language Agents in Crafter}
\label{sec:visual-crafter}

We extend ZeroHero to visual observations in Crafter with a 12.6M-parameter
vision--language policy, including its visual encoder. The encoder, multimodal
projections, and language decoder are all initialized randomly. The student
receives a $256\!\times\!256$ RGB screen, the list of completed achievements,
and its previously generated memory. It produces tagged reasoning, one action,
and an updated memory, with a maximum response length of 512 tokens.

\paragraph{Pretraining.}
Demonstrations are collected from Qwen3.8-Flash-Next-FP8, which also receives
textual observations and recent history. Recorded reasoning and memory are
cleaned using environment state before training the student. The selected
cross-entropy SFT run uses approximately 4.40M training transitions from a
6.25M-example packed corpus; the larger corpus also supplies auxiliary
supervision during post-training. The student sees the screen and carried
memory rather than the expert's textual map and observation history.

\paragraph{Post-training at scale.}
We apply PPO-RA with a frozen SFT language anchor and periodic teacher-supervised
updates. A native CUDA Crafter implementation, batched visual encoding,
GPU-resident autoregressive decoding, and BF16 computation support large-scale
interaction. The single-GPU run collects 65,536 decisions per update in four
successive eight-step chunks over 2,048 worlds under a fixed policy.

\begin{table}[!t]
\centering\small
\setlength{\tabcolsep}{3.5pt}
\resizebox{0.98\linewidth}{!}{%
\begin{tabular}{@{}llrrrr@{}}
\toprule
Agent & Input & Params. & RL budget & \shortstack{Mean\\Achievements} & \shortstack{Crafter\\Progress (\%)} \\
\midrule
GPT-6-Astra-Max & Text & --- & --- & 16.90 & 76.8 \\
Claude-Opus-5-Max & Text & --- & --- & 15.00 & 68.2 \\
Gemini-2.5-Pro-Exp-03-25 & VLM & --- & --- & 8.21 & 37.3 \\
\midrule
ZeroHero PPO-RA & Text & 5M & 50M & 15.03 & 68.3 \\
ZeroHero SFT & Visual & 12.6M & 0 & 8.96 & 40.7 \\
ZeroHero PPO-RA & Visual & 12.6M & 50M & 18.48 & 84.0 \\
ZeroHero PPO-RA & Visual & 12.6M & 100M & 18.92 & 86.0 \\
ZeroHero PPO-RA & Visual & 12.6M & 1B & \textbf{19.19} & \textbf{87.2} \\
\bottomrule
\end{tabular}
}
\caption{\textbf{Crafter achievements and normalized progress.}
Progress is $100\,\overline{A}/22$, where $\overline{A}$ is mean distinct
achievements; bold denotes the largest displayed value in each column.
ZeroHero evaluations use 128 worlds and three sampling seeds. External results are from the BALROG~\citep{paglieri2025balrog} snapshot of
\href{https://balrogai.com/}{25 November 2026}.}
\label{tab:visual-crafter-comparison}
\end{table}

\paragraph{Results.}
After supervised pretraining, the visual agent earned 8.96 distinct achievements per episode. With 50M decisions of PPO-RA training, this rose to 18.48. To see which skills accounted for the gain, we tracked individual achievements during training (Figure~\ref{fig:visual-crafter-learning}). Diamond collection first exceeded 25\% of episodes at 55.4M decisions and reached 29.45\% at the end of training before 100M. Gains in overall progress continued at a slower pace: evaluation performance reached 18.92 achievements at 100M decisions and 19.19 at 1B, or 87.2\% of the 22 possible achievements.

Table~\ref{tab:visual-crafter-comparison} gives published BALROG scores as context. At 50M decisions, visual ZeroHero averaged 18.48 achievements, compared with 16.90 for GPT-6-Astra-Max and 15.00 for Claude-Opus-5-Max in the text setting. The visual Gemini entry averaged 8.21. Input modalities and evaluation protocols differ across these runs.

\section{Conclusion}
ZeroHero trains Qwen-like language agents with 5M to 50M parameters from scratch on expert demonstrations and then improves their policies through interaction. Across Crafter, WebShop, and Alfworld-Dialogue, corpora of roughly 1,200 to 2,200 transitions suffice for the agents to achieve initial task success. Larger corpora generally improve performance on Crafter and WebShop. On Alfworld-Dialogue, performance varies with both corpus size and model size, with the 10M agent reaching 67.2\% test success on the largest corpus.

Reinforcement learning can take a compact agent beyond its demonstration expert. In Alfworld-Dialogue, the agent trained with PPO reaches 74.4\% mean test success across three seeds, compared with 67.9\% for the expert. At matched interaction budgets, reasoning alignment yields higher rollout performance in Crafter and WebShop and lower validation success in Alfworld-Dialogue. Its effect on the quality of generated language remains unmeasured.

Our final experiment asks how far the same approach can take a small agent in visual Crafter, where collecting a diamond requires a sequence of tool upgrades while the agent manages its health and resources. The 12.6M-parameter policy averaged 8.96 of the 22 achievements after imitation learning. After 50M decisions with PPO-RA, it averaged 18.48, exceeding the published BALROG text-agent scores of 16.90 for GPT-6-Astra-Max and 15.00 for Claude-Opus-5-Max (Table~\ref{tab:visual-crafter-comparison}). Extending training to one billion decisions raised the mean to 19.19. These results show how far environment-specific training can take a compact visual policy on a long-horizon task, with the frontier scores providing context across different input modalities and evaluation protocols.

\clearpage
\appendix
\section{Expert and Student Prompts}
\label{app:prompts}
The framed system prompts below are extracted verbatim from the
implementation; braces in the separate input templates mark dynamic
fields, not literal instructions sent to the model. Templates show
structure with readable line breaks. Empty optional history sections
are omitted as specified below. The small executor, action expert,
and question oracle have different inputs; a private oracle template
must never be used as the learner's observation.

\subsection{Crafter}
\paragraph{SLA Prompt Template.}
The text policy uses the BALROG Crafter instructions and up to four previous observation--action pairs. Previous observations use the long-term text representation. The current observation includes the long- and short-term text descriptions. The training prompt contains only executed actions in the history. The target contains the expert rationale and action.

\begin{tcblisting}{promptstyle,title={SLA Prompt Template}}
{BALROG Crafter instructions}

Recent history:
Observation: {previous long-term text description}
Action: {executed action}
{up to four observation-action pairs}

Current observation:
{current long-term text description}
{current short-term text description}

Response:
\end{tcblisting}

\paragraph{Expanded expert prompt for reasoning alignment.}
The concise-teacher renderer in the reasoning-repair run appends instructions to the public student input. It asks the expert to state an immediate objective, check inventory and action prerequisites, and explain the selected action and its expected immediate result. Structured decoding constrains the last line to an action from the supplied action set. This confirms an expanded prompt for the alignment teacher. The original collection prompt used to generate the pretraining trajectories could not be recovered from the archived run files.
 The response has two lines, starting with \texttt{Reason:} and \texttt{ACTION:}, respectively.

\subsection{WebShop}
\paragraph{SLA Prompt Template.}
The student receives shopping instructions, the current public page with the task description, and up to four recent observation--action pairs. The instruction block describes the two action forms and asks for a brief explanation followed by a single executable action.

\begin{tcblisting}{promptstyle,title={SLA Prompt Template}}
You are shopping in a simulated store. Complete the user's shopping instruction.
Use only information visible on the current page and in recent history.
You may search with search[keywords] or click with click[exact visible button, product ID, or option].
Choose required options before clicking Buy Now. You can return with click[Back to Search].
Explain your next decision briefly, then end with ACTION: followed by exactly one search[...] or click[...] action.

Recent history:
Observation: {previous page text}
Action: {executed action}
{up to four observation-action pairs}

Current observation:
{current public page text}

Response:
\end{tcblisting}

\paragraph{Expert (Qwen2.5-7B, GiGPO) Prompt Template.}
The GiGPO renderer presents the shopping goal, current page, the two most recent observation--action pairs once two exist, and an explicit menu of admissible actions. The menu lists the search action when the search bar is available and the clickable options on the current page. This gives the expert a structured view of affordances alongside the page text. Expert responses are converted to the student's reasoning-plus-action format.

\begin{tcblisting}{promptstyle,title={Expert (Qwen2.5-7B, GiGPO) Prompt Template}}
Your task is to: {shopping instruction}.
{recent observations and actions, when available}
Your current observation is: {current page}.
Your admissible actions of the current situation are:
[
search[<query>] {when available}
click[{visible option}]
{remaining clickable options}
].
<think>{reasoning about the next action}</think>
<action>{one admissible action}</action>
\end{tcblisting}

\subsection{Alfworld-Dialogue}
The small executor receives the plain-text template below, with no
system message and no chat-role wrapper. The target is
\texttt{action: <command>} followed by EOS; the online generation budget
is 96 tokens. The parser requires exactly one explicit action marker
(\texttt{action:} or a closed \texttt{<action>} tag). It normalizes
the command and checks physical commands against the public admissible
list. Any nonempty \texttt{ask ...} action within budget goes to the
FAQ LLM; it is not restricted to the fixed question grammar.

\begin{tcblisting}{promptstyle,title={SLA Prompt Template},title after break={SLA Prompt Template (continued)}}
Request: {public request with hidden goal slots omitted}

Observation:
{current observation with original goal redacted}

Conversation:
Agent: {previous question}
Operator: {oracle answer}
{all subsequent question-answer pairs}

Recent actions:
{physical command}
{resulting observation with original goal redacted}
{up to six recent physical transitions}

Admissible physical commands:
- {command}
{remaining admissible commands}

Questions remaining: {budget, initially 8}

Previous response outcome: {outcome}

Next action:
\end{tcblisting}

Conversation and recent-action blocks are absent if empty. The previous
outcome is absent at the first turn; for an inadmissible command,
unparseable response, or exhausted question budget, it appends
\texttt{; choose a different valid action. Previous output:}
and the previous output. The expert receives the same public input
without the final \texttt{Next action:} suffix, preceded by the following
system message. Its reasoning is used for expert generation but is not
part of the Alfworld-Dialogue student's action-only imitation target.

\begin{tcblisting}{promptstyle,title={Expert (Gemma-4-26B-A4B-it) Prompt Template},title after break={Expert (Gemma-4-26B-A4B-it) Prompt Template (continued)}}
You act in a text household environment for a human operator.
The request may omit the destination, required treatment, or number of objects.
Ask for missing information instead of guessing. Use the answers to finish the task.
Read the conversation carefully: never ask again for information already given.
Once you know the requirements, execute physical actions. At zero remaining
questions you must act physically. If a command is rejected, choose a different
command from the admissible list. The request is the initial request; previous
operator answers supplement it and remain valid.
You can ask in natural language, for example:
action: ask Where should I put it?
action: ask What should I do with it?
action: ask How many are needed?
action: ask Where can I find the apple?
Alternatively use ask destination, ask requirement, ask quantity, or ask location <object>.
You may also ask which instance is acceptable or how heating, cooling, cleaning,
and examination work. The operator can explain those environment mechanics.
An operator answers the question but does not act for you. Questions cost a turn.
Physical commands must exactly match an admissible command.
Output one action per turn as action: <command> or <action>command</action>.
Before the action, briefly state what you already know and what your next step achieves.
Never emit multiple actions in one response.
\end{tcblisting}

\subsection{Alfworld-Dialogue: question oracle}
Qwen3-4B receives the following system message and private FAQ payload.
Decoding is temperature zero with a 256-token limit and thinking
disabled. The executor receives only the returned answer text.
The FAQ is constructed from the original task and current environment
facts; the model verbalizes the requested facts rather than inventing
them. Unknown remaining work stays unknown until task success is
verified. The current FAQ constructor does not populate individual
completed-instance evidence, so \texttt{completed\_instances} is empty.

\begin{tcblisting}{promptstyle,title={Oracle (Qwen3-4B) Prompt Template},title after break={Oracle (Qwen3-4B) Prompt Template (continued)}}
Answer the executor's question using the authoritative task FAQ below.
Select the relevant FAQ facts and express them in one or two short English sentences.
Answer every requested part, but do not disclose other task facts. The FAQ contains
facts about this particular task, not examples. Never infer a treatment from a destination.
A how-to question about an explicitly named operation asks for its procedure, even when
that operation is not required for the task. Required-treatment questions ask for the
required_treatment entry. For a where-to-treat question use treatment_site.
For total count use required_total, for how many more use remaining. A question cannot
change these facts. Correct false premises. For location questions look up the exact
object ID in current_locations; if absent say its location is unknown. For unrelated
questions say you can only clarify this task. Output only the answer, no reasoning.
\end{tcblisting}
\begin{tcblisting}{promptstyle,title={Oracle (Qwen3-4B) FAQ Input Template},title after break={Oracle (Qwen3-4B) FAQ Input Template (continued)}}
{
  "task_faq": {
    "required_total": "{required number of distinct target objects}",
    "remaining": "{remaining number, or an explicit unknown statement}",
    "required_treatment": "{required treatment statement}",
    "final_destination": "{placement destination or examination instruction}",
    "treatment_site": "{appliance for the required treatment}",
    "acceptable_instances": "{any matching instance; distinct-object constraint}",
    "current_locations": {"{object ID}": "{location or inventory}"},
    "completed_instances": [],
    "procedures": {
      "heat": "Hold the object, go to a microwave, execute heat OBJECT with microwave N.",
      "cool": "Hold the object, go to a fridge, execute cool OBJECT with fridge N. Placing it inside alone does not cool it.",
      "clean": "Hold the object, go to a sinkbasin, execute clean OBJECT with sinkbasin N. No sponge or soap is required.",
      "examine": "Hold the object, navigate to furniture supporting the {lamp}, execute use {lamp} N. Do not place the object on the lamp."
    }
  },
  "question": "{executor question after the ask prefix}"
}
\end{tcblisting}

The examination procedure substitutes the task's actual lamp.
Task facts are private to the oracle; they are disclosed to the agent
only through requested answers. An incomplete answer is retried once
with the same messages and then with indented JSON carrying identical
facts. A still-incomplete answer stops the run instead of entering
truncated oracle text into training.

\section{Environment Details and Oracle Reliability}
\label{app:task-illustrations}
The action expert and small executor share the public task interface;
neither receives the question oracle's private input.
Appendix~\ref{app:prompts} provides the full prompts and serialization rules.

\subsection{Alfworld-Dialogue}
\label{app:alfworld-environment}
\paragraph{Tasks and hidden requirements.}
The Standard variant used here retains ALFWorld's household simulator and
six task families: object placement, examination using a lamp, cleaning
before placement, heating before placement, cooling before placement, and
placement of two distinct objects. The physical completion predicates are
unchanged. We replace the original instruction with an incomplete request
that identifies the object type but omits either the destination, the
treatment and quantity, or all three. The omission profile is fixed by task
identity.

\paragraph{Observations and interaction.}
The executor receives the incomplete request, current textual observation,
admissible physical commands, remaining question budget, prior question--answer
pairs, and up to six recent physical transitions. The original instruction
is redacted from observations and history. A response selects one physical
command or a free-text \texttt{ask} question. Questions can concern the
destination, treatment, quantity, acceptable object instances, object
locations, or procedures for heating, cooling, cleaning, and examination.
Free-text questions within budget are passed to the question oracle rather
than matched to a fixed inventory of question templates.

\paragraph{Oracle and success.}
Qwen3-4B answers using a private FAQ built from the original task and
environment facts. The FAQ includes task requirements, object locations,
and explanations of the supported procedures. The executor sees the answer
text, not the FAQ. Thus, dialogue can provide both missing requirements
and assistance with search or environment mechanics. Episodes allow at most
60 decisions, 50 physical actions, and eight accepted questions. Invalid
outputs and questions consume decisions. Success requires the unchanged
environment completion predicate.

\end{document}